%% file: main.tex
\documentclass[pmlr]{jmlr}

\usepackage{longtable}
\usepackage{microtype}
 \usepackage{array,multirow,graphicx}
 \usepackage{float}
\usepackage{pgfplots}
\usepackage{multirow}
\usepackage{caption}
\usepackage{subcaption}
\usepackage{booktabs}
\usepackage[load-configurations=version-1]{siunitx} 

 \makeatletter
\def\set@curr@file#1{\def\@curr@file{#1}} 
\makeatother

\renewcommand{\vec}[1]{\boldsymbol{#1}}
\theorembodyfont{\upshape}
\theoremheaderfont{\scshape}
\theorempostheader{:}
\theoremsep{\newline}

\title[Knowledge Base Completion for Constructing Problem-Oriented Medical Records]{Knowledge Base Completion for Constructing Problem-Oriented Medical Records}

\author{\Name{James Mullenbach} \Email{jmullenbach@asapp.com} \\
       \addr ASAPP, New York, NY, USA
       \AND
       \Name{Jordan Swartz} \Email{jordanlswartz@gmail.com} \\
    \vspace{-4mm}
       \AND
       \Name{T. Greg McKelvey} \Email{gmckelvey@asapp.com} \\
       \addr ASAPP, New York, NY, USA
       \AND
       \Name{Hui Dai} \Email{hui.dai@gmail.com} \\
       \addr ASAPP, New York, NY, USA
       \AND
       \Name{David Sontag} \Email{dsontag@asapp.com} \\
       \addr ASAPP, Massachusetts Institute of Technology
       } 

\begin{document}

\maketitle

\begin{abstract}

  Both electronic health records and personal health records are typically organized by data type, with medical problems, medications, procedures, and laboratory results chronologically sorted in separate areas of the chart. As a result, it can be difficult to find all of the relevant information for answering a clinical question about a given medical problem. A promising alternative is to instead organize by problems, with related medications, procedures, and other pertinent information all grouped together. A recent effort by \citet{buchanan2017accelerating} manually defined, through expert consensus, 11 medical problems and the relevant labs and medications for each. We show how to use machine learning on electronic health records to instead automatically construct these problem-based groupings of relevant medications, procedures, and laboratory tests. We formulate the learning task as one of knowledge base completion, and annotate a dataset that expands the set of problems from 11 to 32. We develop a model architecture that exploits both pre-trained concept embeddings and usage data relating the concepts contained in a longitudinal dataset from a large health system. We evaluate our algorithms’ ability to suggest relevant medications, procedures, and lab tests, and find that the approach provides feasible suggestions even for problems that are hidden during training. The dataset, along with code to reproduce our results, is available at \url{https://github.com/asappresearch/kbc-pomr}.
\end{abstract}

\section{Introduction}

\begin{figure}[!ht]
  \centering 
  \includegraphics[scale=0.6]{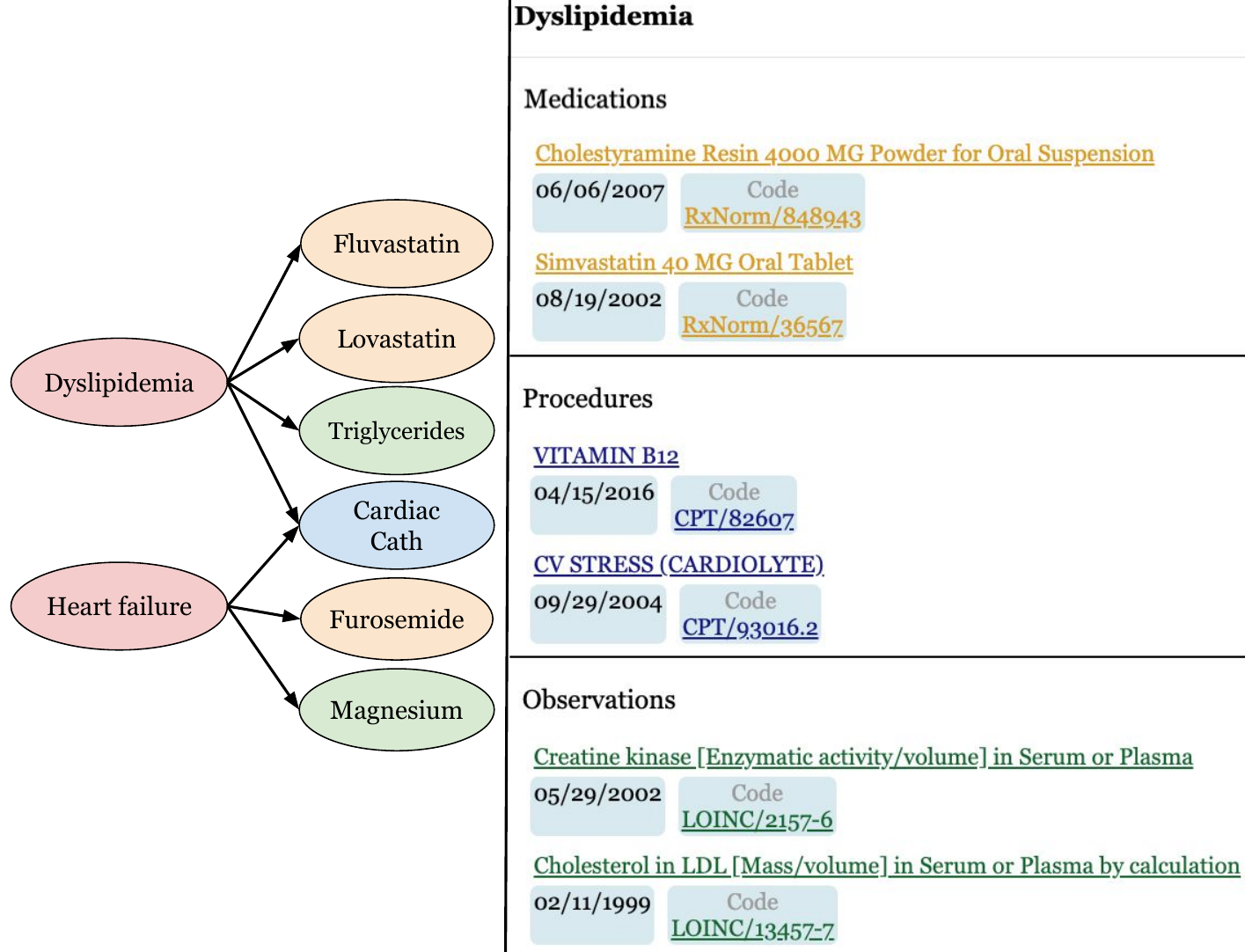}
  \caption{(left) A portion of our annotated knowledge base, as a graph. (right) A mockup of how a POMR might look in practice. The problem, dyslipidemia, is listed as a header, with sections listing relevant data elements of each type. In both figures, gold coloring denotes a medication, blue a procedure, and green a lab result.}
  \label{fig:pomr_mockup}
\end{figure} 

Clinical medicine is a complex task, with each patient representing a unique compilation of health problems involving  overlapping biological systems, complicated care plans, and uncertain states of dynamic progression \citep{kannampallil2011considering}. 
In the electronic health records (EHR) that store patients' data, information about each medical problem may be spread across views that correspond to different data types such as past diagnoses, medications, and procedures \citep{buchanan2017accelerating}. Meanwhile, physicians spend at least as much time interacting with the EHR as they spend interacting with patients \citep{tai2017electronic, sinsky2016allocation}, which may contribute to physician burnout and worse patient outcomes. The problem-oriented medical record (POMR) is a paradigm for presenting medical information that, in contrast to chronological presentations, organizes data around the patient's problem list. This idea predates the widespread use of EHRs and was first introduced by \citet{weed1968medical}, who argued that such an organization would improve physicians' ability to reason about each of their patients' problems. 

To give a motivating example, many patients who suffer from atrial fibrillation, a heart rhythm abnormality, take blood thinners to reduce the likelihood of developing a blood clot. In order for a physician to determine whether the dose of the blood thinner is adequate, they would have to first have to review the problem list of the EHR to see that atrial fibrillation is one of the patient's current medical problems, then scan the medication section to determine what dose the patient is on, and finally navigate to the laboratory section to determine the patient's blood clotting parameters, all of which  involves multiple clicks and time. In the POMR model, all relevant information pertaining to atrial fibrillation would be displayed in the same location within the EHR. We illustrate this concept in \autoref{fig:pomr_mockup}.

We create a seed knowledge base to be used for POMR creation by drafting new problem definitions based on real data and having physician annotators select items pertinent to the management of those problems in an emergency department setting. We then describe and evaluate neural network-based models that are trained to suggest new relations in the knowledge base, and demonstrate the effectiveness of our approach. Constructing a POMR can be construed as a two-step process, in which the first step is identifying the problems or phenotypes in a patient's record, and the second is associating the relevant medications, lab results, and other data to those problems. Our work focuses on the second step, and may be used in conjunction with phenotyping algorithms to form an end-to-end POMR creation pipeline. 
\paragraph{Technical Significance}

In this paper, we frame the problem of constructing a problem-oriented view of a medical record as knowledge base completion, where the entities represent problems, medications, procedures, and laboratory results from patients' structured data. The relations connect problems to the other three entity types, and the relation types simply represent the data source of the target entity type (i.e., \texttt{medication}, \texttt{procedure}, or \texttt{lab}). Then the knowledge base can be represented as lists of medications, procedures, and labs for each problem entity. These lists can then be used downstream as a set of rules to organize patient data around the defined problems; for example, an algorithm could identify diagnosis codes that belong to the problem definition, then attach all co-occurring data elements in the lists. Prior work created lists through manual specification from experts \citep{buchanan2017accelerating}, and in this work we enable automatic creation of these lists by tackling the knowledge base completion task with machine learning.

We develop neural network models that adapt pre-trained medical concept embeddings and learn from both an annotated knowledge base as well as a longitudinal dataset of inpatient and outpatient encounters for 10,000 patients from a regional health system. We evaluate not just on randomly masked entries of the knowledge base, but also by holding out and predicting on entire problems, to test our model's ability to generalize to new diseases. 

\paragraph{Clinical Relevance}
 Having a current and comprehensive problem-oriented view into a patient's data may improve clinician efficiency in retrieving relevant information, answering clinical questions, and completing administrative tasks. This can give time back to physicians, who may spend up to two hours per day working in the EHR outside of working hours \citep{sinsky2016allocation}. Moreover, a less usable EHR is correlated with increased risk of burnout \citep{melnick2019association}, and burnout in turn has been associated with poorer quality of care and increased risk of medical error \citep{panagioti2018association}. 
 
 In this work, we conjecture that automatically linking problems to their associated labs, medications, and procedures will be speedier than the otherwise manual process of determining which items belong to a given medical problem, which involves multiple experts coming to agreement, while maintaining an adequate level of accuracy. 
 
\subsection*{Generalizable Insights about Machine Learning in the Context of Healthcare}

Through our annotation effort, we show that physicians with similar specialties, given a similar situation, often agree on which medications, labs, and procedures in a patient's record are relevant to the treatment of the patient's problems, which is an encouraging sign that automatic construction of a POMR is feasible. We also show that with some fine-tuning, medical concept embeddings initially trained in an unsupervised way can be adapted to predict useful concept relations, which can be leveraged to transform a chronological EHR into a problem-oriented medical record. The result that fine-tuning of embeddings trained on co-occurrence data improves performance for this task also demonstrates that co-occurrence patterns alone are not enough to predict which medications (as an example) are relevant to the treatment of which health problem. Our results also show that using publicly available concept embeddings trained over administrative insurance claims data gives competitive performance to embeddings trained with health system EHR data.

\section{Dataset and Task}

\subsection{Data description}

Though our goal and methods are agnostic to the site at which such models may be deployed, we ground our methods in a dataset of longitudinal health records, covering inpatient, outpatient, and emergency department settings from a large regional health system. The data is anonymized and dates are randomly shifted. 
The data is encounter-based, and each encounter may contain sets of diagnosis codes and procedure, medication, and lab orders. Additionally, some of the procedure, medication, and lab order entries list a corresponding diagnosis code, forming a relation between these diagnoses with the other types of data. When such a relation exists between a diagnosis and medication, lab, or procedure, we call it an \textit{explicit relation}.

Diagnosis codes are recorded primarily using an internal coding system, with some usage of ICD-9 and ICD-10 (International Classification of Diseases). We use provided mappings to convert internal codes to ICD-10, or to SNOMED (Systematized Nomenclature of Medicine) or ICD-9 if no mapping to ICD-10 exists for a code. Procedure orders are recorded using a mixture of CPT (Current Procedural Terminology) and internal codes; however, no mapping is available from internal codes to CPT or other systems. Medication orders are recorded using internal codes with a mapping to RxNorm, which we use. Lab results are recorded using LOINC (Logical Observation Identifiers Names and Codes). 

\paragraph{Statistics}

Our dataset has a total of 10,000 patients, with 1.4 million unique encounters, 1.4 million medication orders, 2.0 million procedure orders, and 6.5 million laboratory results. Of the encounters, 1.6\% are in inpatient or emergency department settings, and 51\% of patients have at least one encounter of those types. 

\subsection{Annotation process}

To learn to suggest sets of medications, procedures, and lab tests for a medical problem, we collect a set of annotated triplets (problem, relation, target). One could also simply use the lists created by \cite{buchanan2017accelerating}, but those lists are limited to a set of 11 problems, the codes chosen are not to our knowledge based on real usage data, and we do not know what guidelines were provided to the experts when choosing medication and lab codes. In order to expand the set of problems we may train on while ensuring consistency in the guidelines used to select codes and the codes presented for annotation, we make our own set of annotations. 

\paragraph{Problem definition} For the annotation task, we present for each problem a list of candidate medication, lab, and procedure codes derived from the dataset. First, we expand the set of problems by defining new ones. Defining problems is not strictly necessary, as one could use groupings from the Clinical Classifications Software (CCS) or Diagnosis Related Groups (DRG), but we aimed to ensure our problems had sufficient examples in our dataset. Here we follow \cite{buchanan2017accelerating} in creating a problem definition as a set of diagnosis codes. To define new problems, we presented an annotator, an emergency medicine attending physician, with a list of diagnosis codes ranked by how many unique patients in the dataset had the code at any point in their history, limited to codes that appear in at least 50 patient records. The annotator then browsed the list in this order and defined new problems as they appeared, by assigning a diagnosis code to a new problem definition as appropriate. After this, the annotator expanded the problem definitions by assigning codes from the rest of the list. After this process we have a set of 32 problems, listed in Appendix \ref{sec:app_a}.

\paragraph{Candidate generation} Then, using these problem definitions, we develop lists of candidate codes based on the dataset. To reduce annotator effort, we aim to provide short lists of likely candidates, so we rank candidate codes using an importance score and provide the top 50 of each data type. We compute the following importance score \citep{rotmensch2017learning} between a problem and a medication, procedure, or lab:

\begin{equation}
    IMPT = \log{(p(x_i=1|y_j=1))} - \log{(p(x_i=1|y_j=0))}
    \label{eqn:init_score}
\end{equation}

where $x_i$ is a binary variable denoting the existence of medication, procedure, or lab $i$ occurring in an encounter record with a reported diagnosis code, and $y_j$ is a binary variable denoting the presence of a diagnosis code in the definition of problem $j$ in an encounter record. This importance score captures the increase in likelihood of a medication, procedure, or lab appearing in an encounter record when a given problem is also recorded in that record. 

To expand beyond this initial set of suggestions, we also perform a second round of annotation using a model trained on the first set. We replace the importance score with the score given by the model in \autoref{eqn:scoring} (see \autoref{sec:distmult}), and present the top 20 suggestions for each (problem, relation) pair.

\paragraph{Guidelines} In initial investigations, we found that a coherent framing led to improved inter-annotator agreement. So we have annotators score each (problem, candidate) pair with a 0 or 1, with the following guidelines: ``1 means this item would be of interest to an emergency medicine physician for the vast majority of cases, when seeing a patient with the condition for the first time.
0 means the item is rarely of interest for this condition'' This framing matches our intended use case of providing a POMR to more efficiently on-board a care provider to a new patient, and matches our annotators' specialty of emergency medicine. To calculate inter-annotator agreement, we had two annotators independently provide labels for a subset of the codes suggested first round of annotations. There were 100 codes, and the resulting Cohen's Kappa score was 0.847, demonstrating the feasibility of achieving high agreement for this task. 

Our final dataset consists of 1,740 positive and 5,024 negative annotations across 32 health problems. 

\section{Methods}

\subsection{DistMult Model}\label{sec:distmult}

The task of POMR construction consists of two main steps: identifying the problems, and identifying the data that are relevant to each problem; this work focuses on the latter task. We frame the problem of relevant data suggestion as knowledge base completion, also known as link prediction, wherein one suggests new relations between existing entities in a knowledge base. This task has seen increasing interest since the successful use of low-dimensional embeddings for the entities and relations in the knowledge base. DistMult \citep{yang2015embedding} was a successful early approach that learns embeddings by applying a relation-specific scoring function $g_r$ to each triplet using a three-way dot product:

\begin{align}
    g_{EMB}^{(r)}=(\vec{e_s}, \vec{e_t}) = \sum_i^de_{s_i}e_{r_i}e_{t_i}
\label{eqn:scoring}
\end{align}

where $\vec{e_s}$ is a source embedding, $\vec{e_t}$ is a target embedding, and $d$ is the dimensionality of embeddings. Based on more recent work \citep{kadlec2017knowledge} in knowledge base completion which showed that DistMult, when well-tuned, is competitive with more sophisticated approaches, we base our model on DistMult. 

\subsection{Initialization and pre-processing}

\paragraph{Pre-trained embeddings}
We use pre-trained embeddings for diagnosis, medication, procedure, and laboratory codes to sensibly initialize the model. Prior work \citep{choi2016learning} learns embeddings for medical concepts from a dataset of insurance claims for 4 million people, exploiting longitudinal aspects of the data. We use the RxNorm, CPT, and LOINC embeddings from this set to initialize the parameters for medication, procedure, and lab codes, respectively, when the embedding code is also present in our dataset. For codes not present in these embeddings, we initialize randomly. To initialize problem embeddings, we combine embeddings for the codes in a problem's definition. We translate any definitions code in SNOMED or ICD-10 to ICD-9, keeping only codes with one-to-one mappings, for use with the embeddings from \cite{choi2016learning}. Embeddings are initialized as the weighted average of each definition code, with weights defined as the frequency of the code in our dataset, normalized across definition codes so the weights sum to 1. Relation embeddings are always initialized to be the all-ones vector, such that the scoring function (\autoref{eqn:scoring}) reduces to a dot product between the source and target embeddings, at the start of training. 

\paragraph{Site-specific embeddings}
We also consider the scenario in which a site trains and uses its own embeddings, which then allows for greater coverage of the codes used and the possibility of usefully including internal codes without mapping. We simulate this scenario by training embeddings on our dataset of 10,000 patients. Specifically, we use Gensim \citep{rehurek_lrec} to train a skip-gram model which treats each encounter as a unit, using the entire set of codes in an encounter as context for a given code. Problem embeddings are initialized using the unweighted average of definition code embeddings. Both sets of embeddings are of dimension 300.

\paragraph{Vocabulary}
The set of RxNorm, LOINC, CPT, and internal codes we use to train models derives from our dataset---it is the set of medication, lab, and procedure codes that occurs at least 5 times. We make this choice to simulate how our globally annotated data would likely be most useful in deployment. The externally trained embeddings are generated from a distinct vocabulary which does not overlap perfectly with our dataset's vocabulary. To ensure a fair comparison, at inference time we evaluate over all codes in the \textit{intersection} of both code sets. Some statistics characterizing the vocabularies and their intersection are in \autoref{tab:vocab_stats}. After translating problem definitions to ICD-9, 98.3\% of definition codes have embeddings from \cite{choi2016learning}. 

\input{mlhc/tables/vocab_stats}

\paragraph{Feature engineering}

Hypothesizing that any POMR implementation is likely to be highly institution-specific, we aim to learn not just from concept-level information, but also from site-level information. We accomplish this by building features from the statistics of our dataset. We count co-occurrences of each (problem,  target) pair in the dataset, and normalize by the count of the target. Each occurrence is counted at most once per patient. We compute these features for varying definitions of co-occurrence. A problem and target are said to co-occur when:

\begin{itemize}
\setlength\itemsep{0em}
    \item An \textit{explicit relation} exists between the two in the data. 
    \item The two appear in the same encounter.
    \item The problem appears within two weeks before or after the target, at the same facility.
    \item The problem appears within two weeks before or after the target, at any facility.
\end{itemize}

We also use data on the specialties of providers listed on encounters. For each problem and target, we count the number of times a provider with a given specialty is listed on an encounter with the problem or target. We use a limited vocabulary of 24 specialties which encompasses 90\% of all encounters that have a provider specialty recorded. Thus for each problem and target we construct a vector of features.

Finally, we use the number of patients for which each problem and target code is recorded as a feature. 

\paragraph{Combining embedding sources with nearest neighbors}

Open-source embeddings such as those from \cite{choi2016learning} are a useful starting point for models, but those models will have limited efficacy on any site-specific codes from an internal vocabulary, and even standardized codes that don't appear in the embeddings' vocabularies. For codes missing from the open-source vocabulary, naively we must randomly initialize embeddings. A more useful initialization for missing codes is to acquire embeddings for them by training on the internal dataset, and exploit their nearest neighbors. Specifically, for each code missing from the external vocabulary, we initialize its embedding by first finding the $k$ nearest neighbors of the code in the internal embedding space, limited to those codes that do exist in the external vocabulary. Then we take the element-wise average of the corresponding \textit{external} embeddings of those neighbor codes, and use that to initialize an embedding for the missing code. We conduct a separate set of experiments using this initialization technique to show its usefulness. 

\subsection{Training and Inference}
\label{sec:training}

\paragraph{Data splits} We divide the annotated triplets at random into training, validation, and test sets at a 70\%, 15\%, 15\% share of the data, respectively. To show that our models are capable of generalizing to new problems, we also create separate train, validation, and test sets, where the test set and validation set consist of all triplets for 5 randomly chosen problems each, and the training set consists of all triplets for the remaining 22 problems. 

\paragraph{Training} After initializing, we train the DistMult model using the ranking loss from the original paper \citep{yang2015embedding}, which guides the model to rank true triplets higher than randomly sampled negative triplets, with a margin. 
However, rather than sample at random from the vocabulary, as is common in knowledge base completion, our dataset has the benefit of having explicit negative examples that result from the annotation process, which we use for training. This has the potential benefit of providing negatives that are still ranked highly according to an importance score, which may improve learning. We shuffle the training set so that during training, each batch consists of a random selection of positive and negative examples.
The model is implemented in PyTorch \citep{paszke2019pytorch} and trained with the Adam optimizer \citep{kingma2014adam}. Learning rate and batch size are tuned by pilot experiments using the validation set.

\paragraph{Using engineered features}

When we use the features we construct from our dataset, we must alter the scoring function. In addition to computing the embedding-based score in \autoref{eqn:scoring}, the model also computes a similar bilinear term to combine the specialty feature vectors for problem and target, using a separate set of relational parameters.

\begin{align}
    g^{(r)}_{SPEC} = (\vec{v}_s,\vec{v}_t) = \sum_i^dv_{s_i}v_{r_i}v_{t_i}
\end{align}

The relational parameters $v_{r_i}$ are similarly initialized to all ones. The other features $\vec{f}(s,t)$ are concatenated with the scores from embeddings and specialty feature vectors, and a final linear layer is used to provide a final score $g^{(r)}$:

\begin{align}
    g^{(r)}(s,t) = \vec{\theta}^\top\left[g^{(r)}_{EMB}(s,t)\oplus g^{(r)}_{SPEC}(s,t) \oplus \vec{f}(s,t)\right]
\end{align}
 
 where $\oplus$ denotes concatenation. 

\paragraph{Inference and Metrics} 
At inference time, for each (source, relation, target) triplet in the validation or test set, we score it along with all negative triplets in the validation/test set with the same problem and relation type. For metrics, we evaluate the mean rank (MR) of the true triplet among the set, along with the mean reciprocal rank (MRR), Hits @ 1 (H@1; frequency of true triplet appearing in top 1), and Hits @ 5 (H@5).  During training, we use early stopping using MR on the validation set to prevent overfitting. 

\subsection{Ontology Baselines}\label{sec:ontology}

Medical ontologies such as the National Drug File Reference Terminology (NDF-RT) contain curated knowledge that can also be leveraged to help construct a POMR. We compare our learning-based methods to rule-based methods that leverage NDF-RT and CPT ontologies to infer medications and procedures, respectively, which are relevant to a problem. Though these ontologies are incomplete and inevitably miss internal codes, having a quantitative comparison can be illuminating. Implementation details on these baselines are described in Appendix \ref{sec:app_b}.

\section{Results}

\input{mlhc/tables/results_annotated_triplets}

\subsection{Held-out triplets}
We first aim to validate our approach by gauging our performance when predicting on randomly held-out triplets. We note that for the held-out problems evaluations, the set of negatives includes all annotated negatives for a given problem and relation type, but in the held-out triplets evaluation, this set is a smaller, random sample of all negatives for the problem-relation type pair. This leads to results that are not directly comparable to held-out problems results.

Results for the held-out triplets evaluation are in \autoref{tab:anno_triplets}. We first compare the two approaches to initializing the model - the externally trained embeddings from \cite{choi2016learning}, and the site-specific embeddings which we trained. To evaluate the impact of learning, we also report results using each set of embeddings as initialized without any training (\textsc{Frozen}). In this evaluation, site-specific embeddings lag the externally trained embeddings when frozen. 

We perform two ablations to determine the source of performance gains from training. By freezing the relation and target embeddings and updating only the problem embedding parameters (\textsc{ProblemOnly}) during training, we evaluate how much performance comes from improved problem representations over their heuristic initializations. Here, the gain for site-specific embeddings is larger than the gain for external embeddings, perhaps suggesting that having more data to pre-train embeddings can lead to diagnosis code embeddings that are robust enough to combine linearly into a useful problem representation. For site-specific embeddings, we see that the list of the 5 most similar problems changes notably before and after training. For atrial fibrillation, before training, the nearest problems are \texttt{[heart failure, arthritis, sleep apnea, asthma, dyslipidemia]}. After training, they are \texttt{[syncope, coronary artery disease, heart failure, hypertension, dyslipidemia]}. This indicates closer alignment to clinical similarity. For arthritis, before training, the neighbor problems are \texttt{[dyslipidemia, GERD, back pain, gout, hypertension]}. After training, they are \texttt{[back pain, rheumatoid arthritis, gout, kidney stone, and headache]}.

By training only the relation embedding parameters (\textsc{RelationOnly}), we can see how much performance derives from separately modeling the relevance of medications, labs, and procedures. We notice that the relation parameters do contribute to performance in terms of MRR, and by combining these gains with those from updating problem embeddings and target embeddings we can improve upon both ablations. Finally, we evaluate adding in our engineered features. This gives a strong boost to performance overall and to nearly all per-type metrics, also narrowing the gap in performance between models that use site-specific and external embeddings.

\input{mlhc/tables/results_annotated_problems}

\subsection{Held-out problems}\label{sec:held_out_problems}

To truly match our intended use case, we evaluate on problems that the model has not encountered during training. These results are in \autoref{tab:annotated_problems}. 

Again, we first compare \cite{choi2016learning}'s embeddings with our site-specific ones without further training. Again in this case, the externally trained embeddings provide better performance, emphasizing their usefulness in the absence of labeled data. We also see that directly applying embeddings in this way is more effective than relying on ontologies. The relatively weak performance of the NDF-RT heuristic on medications in particular seems to be due in large part to the brittleness of the code matching heuristic. Although 65\% of test set medication codes have at least one associated diagnosis code, only 19\% of test set medication codes ultimately end up matching to one of the defined problems. 

We perform a similar set of ablation experiments on this evaluation, however we replace the \textsc{ProblemOnly} ablation with \textsc{Relation+Target}, in which we train just those sets of parameters and keep problems frozen. We make this change because we do not expect to see gains from \textsc{ProblemOnly} training when the validation set contains only problems not seen at train time. Despite this trait of the evaluation, when using \cite{choi2016learning} embeddings we see an improvement in performance in nearly all metrics from \textsc{Relation+Target} training to all embeddings being trained. This gain could be because allowing the problem embeddings to be updated broadens the search space for relation and target parameters that the optimizer will explore. 

Finally, we evaluate the addition of our engineered features, and the results show the usefulness of exploiting auxiliary information for new problems. Again, we see that with all embeddings trained, and using engineered features, the performance gap between models using site-specific and external embeddings becomes much smaller. An initial ablation analysis into the engineered features showed that of those features, most of the gains over models that do not use them came from the explicit relation-based feature. 

\paragraph{Performance by problem}
Some medical problems are more strongly related with some types of entites. For example, urinary tract infection (UTI) is strongly associated with urinalysis and particular antibiotic medications, but there is not a routinely performed procedure for this common condition. In \autoref{tab:matrix} we break down performance on the test set by problem, to help analyze results in this light. Poor performance in sleep apnea medications may not be important, as there are few medications that directly treat that problem. In designing the POMR, it is not expected that every problem would always have associated labs, medications, and procedures, and performing this analysis could be used to decide which suggested elements to turn on. 

\paragraph{Performance by frequency}

\input{mlhc/tables/matrix}

\begin{figure}[!ht]
\centering
    \begin{tabular}{p{7cm}p{7cm}}
        \includegraphics[scale=0.6]{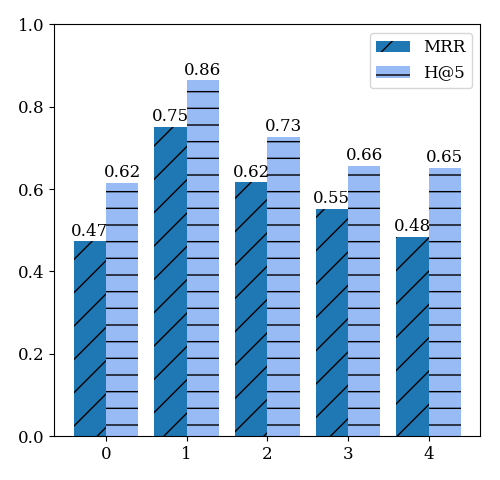} & 
    \includegraphics[scale=0.6]{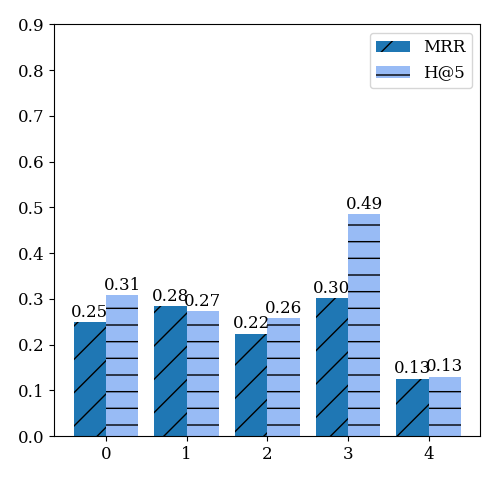}
  \end{tabular}
  \caption{Performance of \cite{choi2016learning} \textsc{+ Features} on held-out problems, grouped by the log of target occurrence counts in the dataset. Log counts are binned and increasing from left to right. On the left, we train as previously described in \autoref{sec:training}, using annotations for negative examples. On the right, we train using negatives randomly sampled from the vocabulary. The bin sizes are (13, 22, 66, 35, 23).}
  \label{fig:freq_hist}
\end{figure} 

In \autoref{fig:freq_hist} we plot the performance of the best model, \cite{choi2016learning} \textsc{+ Features} for different bins of target code frequency. We take the log of the count of each target code in the test set and group into bins before computing the metrics. We show two plots---one trained using annotated negatives (left), and one using negatives randomly sampled from the vocabulary (right). Perhaps counterintuitively, when training with annotated negatives we observe that the most frequent codes do not afford the best performance. Further analysis shows that performance is worse than average over target codes that appear in negative examples in the training set. A possible explanation is that during training, the model updates the embeddings for negative target codes to score them lower, but the resulting embeddings end up further in space from \textit{all} problem embeddings. More frequent codes are more likely to show up as negative training examples, so performance suffers. To investigate this, we train the same model using the randomly sampled negatives. Though the degraded performance for common target codes still appears, the overall performance is notably degraded as well. We leave further investigations into this phenomenon for future work.

\paragraph{Qualitative examples}

\input{mlhc/tables/example_uti}

We provide a list of top suggestions for the best model on this evaluation in \autoref{tab:example_uti}. We see that many of the highest ranked suggestions are ground truth annotations. In medications, Nystatin is a medication to treat fungal infections, which may be correlated via the common cause of obesity. We also see ``Pelvis US'' (pelvic ultrasound), which may investigate pains that turn out to be a symptom of UTI. Knowing this procedure history would be especially useful for chronic UTI patients. ``Bladder lavage'' and ``Cystoscopy/remove object, simple'' suggestions describe bladder procedures that may be seen in similar contexts as UTI. There are still irrelevant suggestions, however, in this case ``Nebulizer treatments''. Further examples for all test set problems are provided in Appendix \ref{sec:app_c}. 

\input{mlhc/tables/results_knn}

\subsection{Combined embeddings evaluation}

To evaluate our $k$-NN approach to initializing embeddings for out-of-vocabulary codes, we perform a separate evaluation. This is necessary because limiting the set of codes to those that are in both embedding sets, as we did in previous experiments, obviates the need for this approach. In this experiment, we evaluate on held-out problems, and do not limit the vocabulary, so the metrics should not be directly compared to those from previous experiments. Models are initialized using the $k$-NN approach for out-of-vocabulary codes and trained the same way as before. $k$ is selected by experiments on the validation set, and is set to 5. Rankings are generated as before by comparing a given code to the full set of negatives for the same problem and relation type. Results are shown in \autoref{tab:knn}. 

Overall, we find that the $k$-NN approach provides better results in MR and Hits@5 than using either embedding source individually, while the overall effect is weak. After training, performance on out-of-vocabulary codes is better when initializing with \cite{choi2016learning} embeddings and the $k$-NN approach than using the site-specific embeddings themselves. However, this comes at the cost of performance on in-vocabulary codes, which leads to only a weak improvement overall after training. The successful use of embeddings trained on disparate data sources for this application merits further study. 

\section{Related Work} 

Prior literature discusses the potential benefits of Weed's problem-oriented medical record \citep{xu2018returning, buchanan2017accelerating}. One review identified determinants that influence use and implementation of problem lists, the foundation of the POMR, concluding that a sociotechnical approach that considers both organizational and technical needs is important for successful implementation \citep{simons2016determinants}. Notably, \citet{wright2015problem} find in a study of ten sites that problem-oriented charting is a factor leading to more complete problem lists. 

We found that works that approach the technical aspects of the problem tend to focus on creating the problem list, without orienting the rest of a patient's data around those problems. For example, \citet{devarakonda2017automated} create problem lists by extracting features from clinical notes and structured data and training decision-tree ensembles in a multilabel way. This work could be considered as part of the larger phenotyping literature, which seeks to usefully classify patients having certain characteristics, one of which may be active problems. Many techniques for phenotyping exist \cite{banda2018advances,ho2014limestone,halpern2016electronic}, which can be combined with our approach to form a POMR without having problem list entries. We found one other work that directly tackles organization of data around medical problems; \cite{juarez2011computing} takes an expert systems approach and orients data around acute problems using temporal constraints. Much previous work \citep{choi2016learning, beam2020clinical, finlayson2014building} investigates how to learn representations of medical concepts in which related medications, procedures, and diagnoses are close in space. Although \cite{choi2016learning} demonstrate that their trained representations are able to usefully group concepts together in a broad sense, they did not evaluate on the specific goal of POMR creation. 

Another work \citep{rotmensch2017learning} sought to create a knowledge graph for healthcare automatically, using concepts extracted from clinical records and narratives. In this work, given the targeted nature of the knowledge we seek to capture, we instead initialize a knowledge graph manually, and then use models learned from clinical data to expand it. We also successfully exploit context from real data using engineered features.

In knowledge base completion, approaches typically focus on general-purpose knowledge bases. Many neural network-based models have been developed and evaluated \citep{yang2015embedding, kadlec2017knowledge}, typically operating over triplets (source, target, relation), though recent successes also exploit information in the entities' neighborhoods \citep{nathani2019learning} and the global graph structure \citep{pinter2018predicting}. These works have not, to our knowledge, exploited information about how the entities in the knowledge base appear in real-world data, beyond that included in embeddings, as our work does by using engineered features.

\section{Discussion}

We've shown that a knowledge base completion approach building on existing models and engineered features can produce a model that is able to usefully organize patient data around well-defined problems. It outperforms baselines that exploit existing medical knowledge bases such as NDF-RT, and those that directly apply pre-trained embeddings.  Importantly, the approach performs well even for previously unseen health problems, such that scaling beyond the 32 initial problems we define for this proof of concept only entails defining a list of diagnosis codes for a problem. 

A next step towards using this research in deployment would be to expand the set of problems, which could be done automatically, using CCS for example, and then directly run the model to rank the medications, procedures and labs in a patient record, using a threshold score to limit the number of suggestions. Then, each suggested element would be shown alongside the condition. Leftover data elements could then be presented either separately in traditional chronological views, or incorporated into the problems for which they have the highest score. Though we do not evaluate on patient records, we note that the Hits@5 results for each data type in \autoref{tab:annotated_problems} are $>$0.5 for the best model. So the majority of the ground truth relevant codes, when compared to $\approx$70 negative codes each, are ranked in the top 5, which suggests these models may be fruitful for patients with a similar number of unique codes of each type. Alternatively, one could also simply repeat our iterated annotation procedure, by providing suggestions for each problem for experts to accept or reject, to save time and effort over less useful suggestions. 

An auxiliary benefit of our annotated dataset is that it may be useful as a way to evaluate learned concept embeddings, similar to how \cite{choi2016learning} evaluate using relations in NDF-RT. Although our dataset was not made for this purpose, the relations it captures would be desirable features of an embedding space. We also observe that the embeddings from \cite{choi2016learning}, trained on a much larger dataset than the site-specific embeddings, perform better without further training. 

Prior work has shown that problem lists and problem-oriented medical records are desirable and useful in practice; we hope our work shows that POMR construction is feasible. By releasing our annotations and using open-source embeddings, we allow other researchers to reproduce our results from \autoref{tab:anno_triplets} and \autoref{tab:annotated_problems} (except those using site-specific embeddings or the engineered features), and hope this allows for further investigation into POMR construction. 

\paragraph{Limitations}

Our approach is a first attempt at this problem and thus has some limitations. First, the usefulness of this model on organizing patient data around a problem relies on that problem having a definition which is strictly coded as a list of diagnosis codes, when in reality the phenotype of diseases are much more complex. However, one need not initialize a problem representation as a linear combination of diagnosis code representations as we do. In general, the adoption of POMR should consider both technical and organizational needs, as pointed out by \cite{simons2016determinants}. Specifically, one would have to consider, for example, the reliability and accuracy of current problem lists, the ability to re-train physicians to use a new system for accessing patient data, and integration concerns such as using institution-internal vocabularies. Only the last of these is partially tackled in this work. 

Additionally, this dataset was collected with the emergency medicine physician's perspective in mind, and what is deemed relevant will certainly vary across medical specialties. The relevance of certain data points may also vary from patient to patient, and even from visit to visit, so this work is most useful when limited to typical patients at initial appointments. 

Finally, the success of this approach ultimately rests on a targeted evaluation that measures whether structuring patient data in this way allows clinicians to answer questions about a patient's medical problems more quickly and accurately than the chronological view. We hope to perform such an evaluation in future work. 

\acks{We thank Christian Koziatek and Rich White for providing the annotations for this project. Yada Pruksachatkun and Sean Adler gave helpful feedback on early drafts of the paper.}

\bibliography{citations.bib}

\appendix
\section{Defined Problems}
\label{sec:app_a}

\input{mlhc/tables/all_problems}

See \autoref{tab:all_prob} for the list of all defined problems.

\section{Ontology baseline implementation}
\label{sec:app_b}

\paragraph{Using NDF-RT}
First, we convert every RxNorm medication code in the test set to a code in NDF-RT. Then, for each NDF-RT medication code we create a set of related NDF-RT diagnosis codes using the ``May Treat'' and ``May Prevent'' relations contained in NDF-RT. These NDF-RT diagnosis codes are then converted to SNOMED when possible, so that each medication code has a list of SNOMED diagnosis codes associated with it. Finally, we augment the list of SNOMED codes to include equivalent ICD-10 and ICD-9 codes. To determine if a medication is relevant to a problem, we look for a match between the problem's definition codes and the medication's associated diagnosis codes. 75\% of test set medication codes are coded in RxNorm, of which 91\% have equivalent NDF-RT codes, of which 95\% have at least one ``May Treat'' or ``May Prevent'' relation. All triplets in the validation set are given a score of 1 if they are deemed relevant by this baseline, and 0 otherwise. Since there are ties with this scoring function, we choose a rank by picking the median position of the example in the set of examples with the same score. For example, suppose there are 10 total triplets in an evaluation step, 1 positive and 9 negative. Suppose 4 negative triplets are deemed relevant and 5 are not. If the positive triplet is relevant, it gets rank 3. If it is not relevant, it gets rank 7.5.

\paragraph{Using CPT}
We could not find a relation analogous to the ``May Treat'' and ``May Prevent'' relations in NDF-RT that would connect procedure concepts to diagnosis codes; however, we use a proxy based on the hierarchy of both CPT and ICD codes. We had a physician provide annotations connecting which parts of the CPT and ICD hierarchy correspond to the same medical disciplines. For example, CPT hierarchy code \texttt{1006056} ``Surgical Procedures on the Cardiovascular System'' falls into the same discipline as ICD-10 code \texttt{I0-I99}, ``Diseases of the circulatory system''. Then, we use this to apply the heuristic that a procedure is relevant to a problem if the CPT's parent code is under the same discipline as any of the problem's definition ICD codes. 

\section{Test set problem examples}
\label{sec:app_c}

Tables \ref{tab:example_htn}, \ref{tab:example_hypokalemia}, \ref{tab:example_thrombo}, and \ref{tab:example_sleep_apnea} contain examples suggestions for all test set problems.

\input{mlhc/tables/example_htn}
\input{mlhc/tables/example_hypokalemia}
\input{mlhc/tables/example_thrombocytopenia}
\input{mlhc/tables/example_sleep_apnea}

\end{document}

%% file: mlhc/tables/vocab_stats.tex
\begin{table*}[]
\centering
\begin{tabular}{@{}lll@{}}
\toprule
Vocab                                    & Statistic                                  & Value  \\ \midrule
Site-specific                            & \# Medication codes                        & 2,294  \\
Site-specific                            & \# Procedure codes                         & 2,655  \\
Site-specific                            & \# Lab codes                               & 1,335  \\
\cite{choi2016learning} & \# Medication (RxNorm) codes               & 802    \\
\cite{choi2016learning} & \# Procedure (CPT) codes                   & 11,746 \\
\cite{choi2016learning} & \# Lab (LOINC) codes                       & 3,093  \\
Intersection                             & Fraction of site-specific medication codes & 25.8\% \\
Intersection                             & Fraction of site-specific procedure codes  & 54.8\% \\
Intersection                             & Fraction of site-specific lab codes        & 55.4\% \\ \bottomrule
\end{tabular}
  \caption{Statistics regarding code sets}
  \label{tab:vocab_stats}
\end{table*}

%% file: mlhc/tables/results_annotated_triplets.tex
\begin{table*}[t]
  \setlength{\tabcolsep}{2pt}
  \small
  \centering
  \begin{tabular}{llllllllllll}
    \toprule
    & \multicolumn{2}{c}{\textbf{Overall}} & \multicolumn{2}{c}{\textbf{Medications}} & \multicolumn{2}{c}{\textbf{Procedures}} & \multicolumn{2}{c}{\textbf{Labs}} \\
      \textbf{Model} & \textbf{MR} & \textbf{MRR} & \textbf{MRR} & \textbf{H@1} & \textbf{MRR} & \textbf{H@1} & \textbf{MRR} & \textbf{H@1} \\
    \midrule
    \cite{choi2016learning} \textsc{Frozen} & 1.83 & 0.749 & 0.753 & 0.571 & 0.824 & 0.688 & 0.601 & 0.371 \\
    \cite{choi2016learning} \textsc{ProblemOnly} & 1.67 & 0.803 & 0.785 & 0.631 & 0.884 & 0.797 & 0.697 & 0.543 \\
    \cite{choi2016learning} \textsc{RelationOnly} & 1.86 & 0.773 & 0.749 & 0.560 & 0.867 & 0.766 & 0.658 & 0.543 \\
    \cite{choi2016learning} & 1.56 & 0.826 & 0.814 & 0.690 & 0.892 & 0.812 & 0.736 & 0.571 \\
    \cite{choi2016learning} \textsc{+ Features} & 1.52 & 0.832 & 0.866 & 0.786 & 0.840 & 0.719 & 0.737 & 0.543 \\
    \midrule
    Site-specific \textsc{Frozen} & 2.93 & 0.570 & 0.597 & 0.381 & 0.611 & 0.438 & 0.429 & 0.200 \\
    Site-specific \textsc{ProblemOnly} & 1.97 & 0.782 & 0.821 & 0.726 & 0.764 & 0.641 & 0.719 & 0.600 \\
    Site-specific \textsc{RelationOnly} & 2.34 & 0.677 & 0.719 & 0.583 & 0.675 & 0.484 & 0.579 & 0.371 \\
    Site-specific & 1.90 & 0.760 & 0.791 & 0.667 & 0.766 & 0.641 & 0.673 & 0.457 \\
    Site-specific \textsc{+ Features} & 1.74 & 0.797 & 0.807 & 0.702 & 0.816 & 0.734 & 0.740 & 0.543 \\
    \bottomrule
    
  \end{tabular}
  \caption{Results on held-out triplets. For MR, lower is better, and higher is better for all other metrics. \textsc{Frozen} means the embeddings are used as initialized with no further training. \textsc{RelationOnly} means that only the relation embeddings are updated during training, and \textsc{ProblemOnly} means that only problem embeddings are updated (see \autoref{eqn:scoring}).}
  \label{tab:anno_triplets}
\end{table*}

%% file: mlhc/tables/results_annotated_problems.tex
\begin{table*}[t]
  \setlength{\tabcolsep}{2pt}
  \small
  \centering
  \begin{tabular}{llllllllllll}
    \toprule
    & \multicolumn{2}{c}{\textbf{Overall}} & \multicolumn{2}{c}{\textbf{Medications}} & \multicolumn{2}{c}{\textbf{Procedures}} & \multicolumn{2}{c}{\textbf{Labs}} \\
      \textbf{Model} & \textbf{MR} & \textbf{MRR} & \textbf{MRR} & \textbf{H@5} & \textbf{MRR} & \textbf{H@5} & \textbf{MRR} & \textbf{H@5} \\
    \midrule
    Ontology baseline & & & 0.066 & 0.023 & 0.150 & 0.270 \\
    \midrule
    \cite{choi2016learning} \textsc{Frozen} &  7.4 & 0.302 & 0.268 & 0.534 & 0.432 & 0.629 & 0.262 & 0.706 \\
    \cite{choi2016learning} \textsc{RelationOnly} & 9.6 & 0.386 & 0.274 & 0.562 & 0.487 & 0.686 & 0.477 & 0.725 \\
    \cite{choi2016learning} \textsc{Relation+Target} & 7.2 & 0.387 & 0.359 & 0.493 & 0.430 & 0.771 & 0.396 & 0.745 \\
    \cite{choi2016learning} & 7.1 & 0.392 & 0.375 & 0.493 & 0.451 & 0.800 & 0.377 & 0.765 \\
    \cite{choi2016learning} \textsc{+ Features} & 5.5 & 0.590 & 0.578 & 0.630 & 0.583 & 0.800 & 0.612 & 0.765 \\
    \midrule
    Site-specific \textsc{Frozen} & 18.9 & 0.175 & 0.182 & 0.219 & 0.256 & 0.257 & 0.109 & 0.098 \\
    Site-specific \textsc{RelationOnly} & 19.3 & 0.244 & 0.192 & 0.219 & 0.265 & 0.314 & 0.305 & 0.373 \\
    Site-specific \textsc{Relation+Target} & 14.7 & 0.270 & 0.222 & 0.247 & 0.155 & 0.343 & 0.419 & 0.510 \\
    Site-specific & 14.3 & 0.248 & 0.209 & 0.219 & 0.176 & 0.400 & 0.352 & 0.392 \\
    Site-specific \textsc{+ Features} & 7.8 & 0.460 & 0.414 & 0.329 & 0.373 & 0.143 & 0.584 & 0.510 \\
    \bottomrule
    
  \end{tabular}
  \caption{Results on held-out problems. \textsc{Relation+Target} means that relation and target embeddings are updating during training, and problem embeddings are kept frozen. ``Ontology baseline'' combines the results from using NDF-RT and CPT heuristics on the medications and procedures, respectively (see \autoref{sec:ontology}).}
  \label{tab:annotated_problems}
\end{table*}

%% file: mlhc/tables/matrix.tex
\begin{table}[]
\centering
\begin{tabular}{@{}llll@{}}
\toprule
                 & Medication & Procedure & Lab  \\ \midrule
Sleep apnea      & 0.00       & 1.00      & 1.00 \\
Hypokalemia      & 0.13       & 1.00      & 0.40 \\
Thrombocytopenia & 0.33       & 0.75      & 0.50 \\
Hypertension     & 0.75       & 0.83      & 1.00 \\
UTI              & 0.84       & 0.67      & 0.91 \\ \bottomrule
\end{tabular}
\caption{Hits@5 values for held-out problems, by data type. Results shown are for the best model in \autoref{sec:held_out_problems}.}
\label{tab:matrix}
\end{table}

%% file: mlhc/tables/example_uti.tex
\begin{table*}[t]
\begin{tabular}{p{3cm}p{5cm}p{6cm}}
\toprule
\textbf{Medication}                             & \textbf{Procedure}                                                    & \textbf{Lab}                                                                                 \\ \midrule
\textcolor{blue}{\textbf{Phenazopyridine}} & \textcolor{blue}{\textbf{Us-renal}} & \textcolor{blue}{\textbf{Piperacillin+tazobactam [susceptibility]}} \\
\textcolor{blue}{\textbf{Ciprofloxacin}} & Bladder lavage/instillation, simple & \textcolor{blue}{\textbf{Bacteria identified in isolate by culture}} \\
\textcolor{blue}{\textbf{Nitrofurantoin}} & \textcolor{blue}{\textbf{Us - abdomen, complete}} & \textcolor{blue}{\textbf{Bacteria identified in unspecified specimen by culture}} \\
\textcolor{blue}{\textbf{Trimethoprim}} & \textcolor{blue}{\textbf{Post void residual bladder us}} & \textcolor{blue}{\textbf{Streptomycin [susceptibility]}} \\
\textcolor{blue}{\textbf{Sulfamethoxazole}} & \textcolor{blue}{\textbf{Cystoscopy}} & Choriogonadotropin (pregnancy test) [presence] in urine \\
\textcolor{blue}{\textbf{Levofloxacin}} & \textcolor{blue}{\textbf{Bladder cath insertion,temp indwell, simple}} & \textcolor{blue}{\textbf{Ampicillin [susceptibility]}} \\
Nystatin & \textcolor{blue}{\textbf{Cystoscopy/remove object, simple}} & \textcolor{blue}{\textbf{Aztreonam [susceptibility]}} \\
\textcolor{blue}{\textbf{Tamsulosin}} & \textcolor{blue}{\textbf{Abdomen ct w/o iv and with po contrast}} & \textcolor{blue}{\textbf{Cefoxitin [susceptibility]}} \\
\textcolor{blue}{\textbf{Cephalexin}} & \textcolor{blue}{\textbf{Abd/pelvis ct w/ + w/o iv contrast (no po)}} & \textcolor{blue}{\textbf{Cefepime [susceptibility]}} \\
Enoxaparin & \textcolor{blue}{\textbf{Cystoscopy/insertion of stent}} & \textcolor{blue}{\textbf{Trimethoprim+sulfamethoxazole [susceptibility]}} \\
\bottomrule
\end{tabular}
\caption{Example top-10 suggestions from the best-performing model from \autoref{sec:held_out_problems} (\cite{choi2016learning} + features) for the held-out test set problem UTI. Candidates are drawn from the set of all triplets in the test set. Suggestions in blue are positive examples. The medication nitrofurantoin is an antibiotic commonly used to treat a UTI, so knowing whether a patient was previously prescribed it may be important for current treatment due to antibiotic resistance considerations. The procedure "Us-renal" stands for a renal ultrasound, a commonly ordered procedure for patients who frequently have UTI's. Many of the lab tests are susceptibility tests, which are important to know for personalizing antibiotic recommendations.}
\label{tab:example_uti}
\end{table*}

%% file: mlhc/tables/results_knn.tex
\begin{table*}[t]
  \setlength{\tabcolsep}{2pt}
  \small
  \centering
  \begin{tabular}{lccc@{\hskip 0.5cm}cc@{\hskip 0.5cm}cc}
    \toprule
    & \multicolumn{3}{c}{\textbf{Overall}} & \multicolumn{2}{c}{\textbf{IV}} & \multicolumn{2}{c}{\textbf{OOV}} \\
      \textbf{Model} & \textbf{MR} & \textbf{MRR} & \textbf{H@5} & \textbf{MRR} & \textbf{H@5} & \textbf{MRR} & \textbf{H@5} \\
    \midrule
    Site-specific \textsc{Frozen} & 20.1 & 0.163 & 0.182 & 0.175 & 0.189 & 0.134 & 0.167 \\
    \cite{choi2016learning} \textsc{Frozen} &  21.1 & 0.224 & 0.431 & 0.302 & 0.610 & 0.035 & 0.000 \\
    \cite{choi2016learning} \textsc{Frozen} 5-NN &  14.5 & 0.209 & 0.276 & 0.250 & 0.346 & 0.110 & 0.106 \\
    \midrule
    Site-specific & 17.8 & 0.199 & 0.249 & 0.196 & 0.252 & 0.208 & 0.242 \\
    \cite{choi2016learning} &  14.3 & 0.324 & 0.427 & 0.410 & 0.541 & 0.116 & 0.152 \\
    \cite{choi2016learning} 5-NN &  11.7 & 0.318 & 0.471 & 0.321 & 0.465 & 0.312 & 0.485 \\
    \bottomrule
    
  \end{tabular}
  \caption{Results on held-out problems, with no vocab restriction, to evaluate the $k$-NN embedding initialization approach. ``IV'' refers to codes that are \textbf{I}n the \cite{choi2016learning} \textbf{V}ocabulary, while ``OOV'' refers to codes that are \textbf{O}ut \textbf{O}f \textbf{V}ocabulary.}
  \label{tab:knn}
\end{table*}

%% file: mlhc/tables/all_problems.tex
\begin{table}[]
\begin{tabular}{|c|}
\toprule
\textbf{Anemia}                                \\
Arthritis                                      \\
Asthma                                         \\
Atrial fibrillation                            \\
Back pain                                      \\
Cholelithiasis                                 \\
\textbf{Chronic kidney disease}                \\
\textbf{Chronic obstructive pulmonary disease} \\
\textbf{Coronary artery disease}               \\
Cough                                          \\
Dermatitis                                     \\
\textbf{Diabetes}                              \\
Diverticulosis/Diverticulitis                  \\
\textbf{Dyslipidemia}                          \\
Gastroesophageal reflux disease                \\
Gout                                           \\
Headache                                       \\
Heart failure                                  \\
Hematuria                                      \\
\textbf{Hypertension}                          \\
Hypokalemia                                    \\
Kidney stone                                   \\
\textbf{Mood disorders, including depression}  \\
\textbf{Osteoporosis}                          \\
Rheumatoid Arthritis                           \\
\textbf{Seizure disorder}                      \\
Sleep apnea                                    \\
Syncope                                        \\
Thrombocytopenia                               \\
\textbf{Thyroid hormone disorders}             \\
Uterine fibroid                                \\
Urinary tract infection                       \\
\bottomrule
\end{tabular}
\caption{All problems. Those in bold are those initially defined by \citet{buchanan2017accelerating}}.
\label{tab:all_prob}
\end{table}

%% file: mlhc/tables/example_htn.tex
\begin{table*}[t]
\begin{tabular}{lp{5cm}p{5cm}}
\toprule
\textbf{Medication}                             & \textbf{Procedure}                                                    & \textbf{Lab}                                                                                 \\ \midrule
\textcolor{blue}{\textbf{Atenolol}} & \textcolor{blue}{\textbf{Ekg complete (tracing and interp)}} & \textcolor{blue}{\textbf{Creatinine [mass/volume] in serum or plasma}} \\
\textcolor{blue}{\textbf{Diltiazem}} & \textcolor{blue}{\textbf{Ekg (hospital based)}} & Potassium [moles/volume] in serum or plasma \\
\textcolor{blue}{\textbf{Metoprolol}} & Holter complete (comm prac) & Inr in platelet poor plasma by coagulation assay \\
\textcolor{blue}{\textbf{Lisinopril}} & \textcolor{blue}{\textbf{Echo, stress (exercise)}} & Platelets [\#/volume] in blood by automated count \\
\textcolor{blue}{\textbf{Doxazosin}} & Methylprednisolone acetate inj 40 mg & Digoxin [mass/volume] in serum or plasma \\
\textcolor{blue}{\textbf{Nitroglycerin}} & Pulse oximetry sngl determin (op) & Leukocytes [\#/volume] in blood by automated count \\
\textcolor{blue}{\textbf{Olmesartan}} & Mri-brain without contrast & Glucose [mass/volume] in serum or plasma \\
\textcolor{blue}{\textbf{Chlorthalidone}} & Abdomen ct w/o iv and with po contrast & Natriuretic peptide b [mass/volume] in serum or plasma \\
\textcolor{blue}{\textbf{Amlodipine}} & Pacemaker with interpretation & Prothrombin time (pt) \\
\textcolor{blue}{\textbf{Hydrochlorothiazide}} & \textcolor{blue}{\textbf{Echo, complete (2d), trans-thoracic}} & Sodium [moles/volume] in serum or plasma \\
\bottomrule
\end{tabular}
\caption{Example top-10 suggestions from the best-performing model (\cite{choi2016learning} + features) for the held-out test set problem Hypertension. Candidates are drawn from the set of all triplets in the test set. Suggestions in blue are positive examples. Note that for hypertension, there is only 1 positive annotated lab.}
\label{tab:example_htn}
\end{table*}

%% file: mlhc/tables/example_hypokalemia.tex
\begin{table*}[t]
\begin{tabular}{lp{5cm}p{5cm}}
\toprule
\textbf{Medication}                             & \textbf{Procedure}                                                    & \textbf{Lab}                                                                                 \\ \midrule
Metoprolol & \textcolor{blue}{\textbf{Ekg (hospital based)}} & Lipase [enzymatic activity/volume] in serum or plasma \\
Enoxaparin & \textcolor{blue}{\textbf{(ekg) tracing only}} & Natriuretic peptide b [mass/volume] in serum or plasma \\
Labetalol & Dual-lead pacemaker + reprogram & Prothrombin time (pt) \\
Nitroglycerin & Echo, complete (2d), trans-thoracic & \textcolor{blue}{\textbf{Creatinine [mass/volume] in serum or plasma}} \\
Heparin & Holter hook-up (comm prac) & Inr in platelet poor plasma by coagulation assay \\
Lorazepam & Chest ct, no contrast & Bilirubin.total [mass/volume] in serum or plasma \\
Amlodipine & \textcolor{blue}{\textbf{Ekg complete (tracing and interp)}} & Leukocytes [\#/volume] in blood by automated count \\
Carvedilol & Dual-lead defibrillator + reprogram & \textcolor{blue}{\textbf{Potassium [moles/volume] in serum or plasma}} \\
Clopidogrel & Nuc med, myocardial stress - pharmacologic & Creatine kinase.mb [mass/volume] in serum or plasma \\
Oxygen & Defibrillator remote interrogation eval/interp,to 90d & Renin [enzymatic activity/volume] in plasma \\
\bottomrule
\end{tabular}
\caption{Example top-10 suggestions from the best-performing model (\cite{choi2016learning} + features) for the held-out test set problem Hypokalemia. Candidates are drawn from the set of all triplets in the test set. Suggestions in blue are positive examples. }
\label{tab:example_hypokalemia}
\end{table*}

%% file: mlhc/tables/example_thrombocytopenia.tex
\begin{table*}[t]
\begin{tabular}{lp{5cm}p{5cm}}
\toprule
\textbf{Medication}                             & \textbf{Procedure}                                                    & \textbf{Lab}                                                                                 \\ \midrule
Metoprolol & Us - abdomen, complete & Reticulocytes/100 erythrocytes in blood by automated count \\
Prednisone & \textcolor{blue}{\textbf{Filgrastim inj 480 mcg}} & \textcolor{blue}{\textbf{Hemoglobin [mass/volume] in blood}} \\
Sulfamethoxazole & Abdomen ct w/o iv and with po contrast & Creatinine [mass/volume] in serum or plasma \\
Acetaminophen & Us - abdomen,  limited & \textcolor{blue}{\textbf{Inr in platelet poor plasma by coagulation assay}} \\
Cephalexin & \textcolor{blue}{\textbf{Bone marrow biopsy}} & Ferritin [mass/volume] in serum or plasma \\
\textcolor{blue}{\textbf{Pegfilgrastim}} & \textcolor{blue}{\textbf{Bone marrow aspiration}} & Bacteria identified in isolate by culture \\
\textcolor{blue}{\textbf{Ferrous sulfate}} & \textcolor{blue}{\textbf{Filgrastim inj 300 mcg}} & Circulating tumor cells.breast [\#/volume] in blood \\
Enalapril & \textcolor{blue}{\textbf{Ct - head/brain w/o contrast}} & Sodium [moles/volume] in serum or plasma \\
\textcolor{blue}{\textbf{Ibuprofen}} & Alteplase recomb 1mg & Natriuretic peptide b [mass/volume] in serum or plasma \\
Tamsulosin & \textcolor{blue}{\textbf{Cbc w/o plt}} & Leukocytes [\#/volume] in blood by automated count \\
\bottomrule
\end{tabular}
\caption{Example top-10 suggestions from the best-performing model (\cite{choi2016learning} + features) for the held-out test set problem Thrombocytopenia. Candidates are drawn from the set of all triplets in the test set. Suggestions in blue are positive examples. }
\label{tab:example_thrombo}
\end{table*}

%% file: mlhc/tables/example_sleep_apnea.tex
\begin{table*}[t]
\begin{tabular}{lp{5cm}p{5cm}}
\toprule
\textbf{Medication}                             & \textbf{Procedure}                                                    & \textbf{Lab}                                                                                 \\ \midrule
Montelukast & \textcolor{blue}{\textbf{Sleep study, w/ cpap (treatment settings)}} & Natriuretic peptide.b prohormone n-terminal [mass/volume] in serum or plasma \\
Ipratropium & \textcolor{blue}{\textbf{Positive airway pressure (cpap)}} & \textcolor{blue}{\textbf{Natriuretic peptide b [mass/volume] in serum or plasma}} \\
Enoxaparin & \textcolor{blue}{\textbf{Sleep study, w/o cpap}} & Nicotine [mass/volume] in urine \\
Exenatide & Pulse ox w/ rest/exercise, multiple (op) & Inr in platelet poor plasma by coagulation assay \\
Lamotrigine & Basic spirometry & Creatinine [mass/volume] in serum or plasma \\
Oxycodone & (ekg) tracing only & Leukocytes [\#/volume] in blood by automated count \\
Albuterol & Ekg complete (tracing and interp) & Cotinine [mass/volume] in urine \\
Fluticasone & Pneumogram, peds multi-channel & Hemoglobin [mass/volume] in blood \\
Azithromycin & Ekg (hospital based) & Opiates [presence] in urine by screen method \\
Clonazepam & Intervene hlth/behave, indiv & Ferritin [mass/volume] in serum or plasma \\
\bottomrule
\end{tabular}
\caption{Example top-10 suggestions from the best-performing model (\cite{choi2016learning} + features) for the held-out test set problem Sleep apnea. Candidates are drawn from the set of all triplets in the test set. Suggestions in blue are positive examples.}
\label{tab:example_sleep_apnea}
\end{table*}

%% file: main.bbl
\begin{thebibliography}{26}
\providecommand{\natexlab}[1]{#1}
\providecommand{\url}[1]{\texttt{#1}}
\expandafter\ifx\csname urlstyle\endcsname\relax
  \providecommand{\doi}[1]{doi: #1}\else
  \providecommand{\doi}{doi: \begingroup \urlstyle{rm}\Url}\fi

\bibitem[Banda et~al.(2018)Banda, Seneviratne, Hernandez-Boussard, and
  Shah]{banda2018advances}
Juan~M Banda, Martin Seneviratne, Tina Hernandez-Boussard, and Nigam~H Shah.
\newblock Advances in electronic phenotyping: from rule-based definitions to
  machine learning models.
\newblock \emph{Annual review of biomedical data science}, 1:\penalty0 53--68,
  2018.

\bibitem[Beam et~al.(2020)Beam, Kompa, Schmaltz, Fried, Weber, Palmer, Shi,
  Cai, and Kohane]{beam2020clinical}
Andrew~L Beam, Benjamin Kompa, Allen Schmaltz, Inbar Fried, Griffin Weber,
  Nathan Palmer, Xu~Shi, Tianxi Cai, and Isaac~S Kohane.
\newblock Clinical concept embeddings learned from massive sources of
  multimodal medical data.
\newblock In \emph{Pacific Symposium on Biocomputing. Pacific Symposium on
  Biocomputing}, volume~25, page 295. NIH Public Access, 2020.

\bibitem[Buchanan(2017)]{buchanan2017accelerating}
Joel Buchanan.
\newblock Accelerating the benefits of the problem oriented medical record.
\newblock \emph{Applied clinical informatics}, 26\penalty0 (01):\penalty0
  180--190, 2017.

\bibitem[Choi et~al.(2016)Choi, Chiu, and Sontag]{choi2016learning}
Youngduck Choi, Chill Yi-I Chiu, and David Sontag.
\newblock Learning low-dimensional representations of medical concepts.
\newblock \emph{AMIA Summits on Translational Science Proceedings},
  2016:\penalty0 41, 2016.

\bibitem[Devarakonda et~al.(2017)Devarakonda, Mehta, Tsou, Liang, Nowacki, and
  Jelovsek]{devarakonda2017automated}
Murthy~V Devarakonda, Neil Mehta, Ching-Huei Tsou, Jennifer~J Liang, Amy~S
  Nowacki, and John~Eric Jelovsek.
\newblock Automated problem list generation and physicians perspective from a
  pilot study.
\newblock \emph{International journal of medical informatics}, 105:\penalty0
  121--129, 2017.

\bibitem[Finlayson et~al.(2014)Finlayson, LePendu, and
  Shah]{finlayson2014building}
Samuel~G Finlayson, Paea LePendu, and Nigam~H Shah.
\newblock Building the graph of medicine from millions of clinical narratives.
\newblock \emph{Scientific data}, 1:\penalty0 140032, 2014.

\bibitem[Halpern et~al.(2016)Halpern, Horng, Choi, and
  Sontag]{halpern2016electronic}
Yoni Halpern, Steven Horng, Youngduck Choi, and David Sontag.
\newblock Electronic medical record phenotyping using the anchor and learn
  framework.
\newblock \emph{Journal of the American Medical Informatics Association},
  23\penalty0 (4):\penalty0 731--740, 2016.

\bibitem[Ho et~al.(2014)Ho, Ghosh, Steinhubl, Stewart, Denny, Malin, and
  Sun]{ho2014limestone}
Joyce~C Ho, Joydeep Ghosh, Steve~R Steinhubl, Walter~F Stewart, Joshua~C Denny,
  Bradley~A Malin, and Jimeng Sun.
\newblock Limestone: High-throughput candidate phenotype generation via tensor
  factorization.
\newblock \emph{Journal of biomedical informatics}, 52:\penalty0 199--211,
  2014.

\bibitem[Juarez et~al.(2011)Juarez, Campos, Gomariz, and
  Morales]{juarez2011computing}
Jose~M Juarez, Manuel Campos, Antonio Gomariz, and Antonio Morales.
\newblock Computing problem oriented medical records.
\newblock In \emph{International Workshop on Knowledge Representation for
  Health Care}, pages 117--130. Springer, 2011.

\bibitem[Kadlec et~al.(2017)Kadlec, Bajgar, and
  Kleindienst]{kadlec2017knowledge}
Rudolf Kadlec, Ond{\v{r}}ej Bajgar, and Jan Kleindienst.
\newblock Knowledge base completion: Baselines strike back.
\newblock In \emph{Proceedings of the 2nd Workshop on Representation Learning
  for NLP}, pages 69--74, 2017.

\bibitem[Kannampallil et~al.(2011)Kannampallil, Schauer, Cohen, and
  Patel]{kannampallil2011considering}
Thomas~G Kannampallil, Guido~F Schauer, Trevor Cohen, and Vimla~L Patel.
\newblock Considering complexity in healthcare systems.
\newblock \emph{Journal of biomedical informatics}, 44\penalty0 (6):\penalty0
  943--947, 2011.

\bibitem[Kingma and Ba(2015)]{kingma2014adam}
Diederik Kingma and Jimmy Ba.
\newblock Adam: A method for stochastic optimization.
\newblock In \emph{International Conference on Learning Representations}, 2015.

\bibitem[Melnick et~al.(2019)Melnick, Dyrbye, Sinsky, Trockel, West, Nedelec,
  Tutty, and Shanafelt]{melnick2019association}
Edward~R Melnick, Liselotte~N Dyrbye, Christine~A Sinsky, Mickey Trockel,
  Colin~P West, Laurence Nedelec, Michael~A Tutty, and Tait Shanafelt.
\newblock The association between perceived electronic health record usability
  and professional burnout among us physicians.
\newblock In \emph{Mayo Clinic Proceedings}. Elsevier, 2019.

\bibitem[Nathani et~al.(2019)Nathani, Chauhan, Sharma, and
  Kaul]{nathani2019learning}
Deepak Nathani, Jatin Chauhan, Charu Sharma, and Manohar Kaul.
\newblock Learning attention-based embeddings for relation prediction in
  knowledge graphs.
\newblock In \emph{Proceedings of the 57th Annual Meeting of the Association
  for Computational Linguistics}, pages 4710--4723, 2019.

\bibitem[Panagioti et~al.(2018)Panagioti, Geraghty, Johnson, Zhou,
  Panagopoulou, Chew-Graham, Peters, Hodkinson, Riley, and
  Esmail]{panagioti2018association}
Maria Panagioti, Keith Geraghty, Judith Johnson, Anli Zhou, Efharis
  Panagopoulou, Carolyn Chew-Graham, David Peters, Alexander Hodkinson, Ruth
  Riley, and Aneez Esmail.
\newblock Association between physician burnout and patient safety,
  professionalism, and patient satisfaction: a systematic review and
  meta-analysis.
\newblock \emph{JAMA internal medicine}, 178\penalty0 (10):\penalty0
  1317--1331, 2018.

\bibitem[Paszke et~al.(2019)Paszke, Gross, Massa, Lerer, Bradbury, Chanan,
  Killeen, Lin, Gimelshein, Antiga, et~al.]{paszke2019pytorch}
Adam Paszke, Sam Gross, Francisco Massa, Adam Lerer, James Bradbury, Gregory
  Chanan, Trevor Killeen, Zeming Lin, Natalia Gimelshein, Luca Antiga, et~al.
\newblock Pytorch: An imperative style, high-performance deep learning library.
\newblock In \emph{Advances in Neural Information Processing Systems}, pages
  8024--8035, 2019.

\bibitem[Pinter and Eisenstein(2018)]{pinter2018predicting}
Yuval Pinter and Jacob Eisenstein.
\newblock Predicting semantic relations using global graph properties.
\newblock In \emph{Proceedings of the 2018 Conference on Empirical Methods in
  Natural Language Processing}, pages 1741--1751, 2018.

\bibitem[{\v R}eh{\r u}{\v r}ek and Sojka(2010)]{rehurek_lrec}
Radim {\v R}eh{\r u}{\v r}ek and Petr Sojka.
\newblock {Software Framework for Topic Modelling with Large Corpora}.
\newblock In \emph{{Proceedings of the LREC 2010 Workshop on New Challenges for
  NLP Frameworks}}, pages 45--50, Valletta, Malta, May 2010. ELRA.
\newblock \url{http://is.muni.cz/publication/884893/en}.

\bibitem[Rotmensch et~al.(2017)Rotmensch, Halpern, Tlimat, Horng, and
  Sontag]{rotmensch2017learning}
Maya Rotmensch, Yoni Halpern, Abdulhakim Tlimat, Steven Horng, and David
  Sontag.
\newblock Learning a health knowledge graph from electronic medical records.
\newblock \emph{Scientific reports}, 7\penalty0 (1):\penalty0 1--11, 2017.

\bibitem[Simons et~al.(2016)Simons, Cillessen, and
  Hazelzet]{simons2016determinants}
Sereh~MJ Simons, Felix~HJM Cillessen, and Jan~A Hazelzet.
\newblock Determinants of a successful problem list to support the
  implementation of the problem-oriented medical record according to recent
  literature.
\newblock \emph{BMC medical informatics and decision making}, 16\penalty0
  (1):\penalty0 102, 2016.

\bibitem[Sinsky et~al.(2016)Sinsky, Colligan, Li, Prgomet, Reynolds, Goeders,
  Westbrook, Tutty, and Blike]{sinsky2016allocation}
Christine Sinsky, Lacey Colligan, Ling Li, Mirela Prgomet, Sam Reynolds,
  Lindsey Goeders, Johanna Westbrook, Michael Tutty, and George Blike.
\newblock Allocation of physician time in ambulatory practice: a time and
  motion study in 4 specialties.
\newblock \emph{Annals of internal medicine}, 165\penalty0 (11):\penalty0
  753--760, 2016.

\bibitem[Tai-Seale et~al.(2017)Tai-Seale, Olson, Li, Chan, Morikawa, Durbin,
  Wang, and Luft]{tai2017electronic}
Ming Tai-Seale, Cliff~W Olson, Jinnan Li, Albert~S Chan, Criss Morikawa, Meg
  Durbin, Wei Wang, and Harold~S Luft.
\newblock Electronic health record logs indicate that physicians split time
  evenly between seeing patients and desktop medicine.
\newblock \emph{Health Affairs}, 36\penalty0 (4):\penalty0 655--662, 2017.

\bibitem[Weed(1968)]{weed1968medical}
Lawrence~L Weed.
\newblock Medical records that guide and teach (concluded).
\newblock \emph{Yearbook of Medical Informatics}, 212:\penalty0 1, 1968.

\bibitem[Wright et~al.(2015)Wright, McCoy, Hickman, Hilaire, Borbolla,
  Bowes~III, Dixon, Dorr, Krall, Malholtra, et~al.]{wright2015problem}
Adam Wright, Allison~B McCoy, Thu-Trang~T Hickman, Daniel~St Hilaire, Damian
  Borbolla, Watson~A Bowes~III, William~G Dixon, David~A Dorr, Michael Krall,
  Sameer Malholtra, et~al.
\newblock Problem list completeness in electronic health records: a multi-site
  study and assessment of success factors.
\newblock \emph{International journal of medical informatics}, 84\penalty0
  (10):\penalty0 784--790, 2015.

\bibitem[Xu and Papier(2018)]{xu2018returning}
Shuai Xu and Arthur Papier.
\newblock Returning to (electronic) health records that guide and teach.
\newblock \emph{The American journal of medicine}, 131\penalty0 (7):\penalty0
  723--725, 2018.

\bibitem[Yang et~al.(2015)Yang, Yih, He, Gao, and Deng]{yang2015embedding}
Bishan Yang, Scott Wen-tau Yih, Xiaodong He, Jianfeng Gao, and Li~Deng.
\newblock Embedding entities and relations for learning and inference in
  knowledge bases.
\newblock In \emph{Proceedings of the International Conference on Learning
  Representations (ICLR) 2015}, May 2015.
\newblock URL
  \url{https://www.microsoft.com/en-us/research/publication/embedding-entities-and-relations-for-learning-and-inference-in-knowledge-bases/}.

\end{thebibliography}
